\pdfoutput=1
\documentclass[10pt]{article} 
\usepackage[margin=0.5in]{geometry}
\usepackage{float}
\usepackage{times}
\usepackage{amssymb}
\usepackage{sectsty}
\usepackage{url,hyperref}
\usepackage{hyperref}
\usepackage{graphicx}
\usepackage{caption}
\usepackage{subcaption}
\usepackage[font=small,skip=0pt]{caption}
\usepackage{indentfirst}
\hypersetup{
    colorlinks=true,
    linkcolor=black,
    filecolor=black,      
    urlcolor=black,
}

\sectionfont{\fontsize{10}{15}\selectfont}
\subsectionfont{\fontsize{10}{15}\selectfont}
\newcommand{\norm}[1]{\|#1\|}
\begin{document}

\title{A Particle Filtering Framework for Integrity Risk of GNSS-Camera Sensor Fusion}

\author{Adyasha Mohanty, Shubh Gupta and Grace Xingxin Gao\\
\date{}
\textit{Stanford University}\\
}
\maketitle
\section*{BIOGRAPHIES}
Adyasha Mohanty is a graduate student in the Department of Aeronautics and Astronautics at Stanford University. She graduated with a B.S. in Aerospace Engineering from Georgia Institute of Technology in 2019. \par

Shubh Gupta is a graduate student in the Department of Electrical Engineering at Stanford University. He received his B.Tech degree in Electrical Engineering with a minor in Computer Science from the Indian Institute of Technology Kanpur in 2018. 

Grace Xingxin Gao is an assistant professor in the Department of Aeronautics and Astronautics at Stanford University. Before joining Stanford University, she was faculty at University of Illinois at Urbana-Champaign. She obtained her Ph.D. degree at Stanford University. Her research is on robust and secure positioning, navigation and timing with applications to manned and unmanned aerial vehicles, robotics, and power systems. 

\section*{ABSTRACT}
Adopting a joint approach towards state estimation and integrity monitoring results in unbiased integrity monitoring unlike traditional approaches. So far, a joint approach was used in Particle RAIM \cite{p} for GNSS measurements only.
In our work, we extend Particle RAIM to a GNSS-camera fused system for joint state estimation and integrity monitoring.
To account for vision faults, we derive a probability distribution over position from camera images using map-matching.
We formulate a Kullback-Leibler Divergence \cite{kl} metric to assess the consistency of GNSS and camera measurements and mitigate faults during sensor fusion.
The derived integrity risk upper bounds the probability of Hazardously Misleading Information (HMI).
Experimental validation on a real-world dataset shows that our algorithm produces less than 11 m position error and the integrity risk over bounds the probability of HMI with 0.11 failure rate for an 8 m Alert Limit in an urban scenario.

\section{INTRODUCTION}
In urban environments, GNSS signals suffer from lack of continuous satellite signal availability, non line-of-sight (NLOS) errors and multi-path effects.
Thus, it is important to quantify the integrity or measure of trust in the correctness of the positioning solution provided by the navigation system.
Traditional integrity monitoring approaches \cite{trad} provide point positioning estimates i.e. the state estimation algorithm is assumed to be correct and then the integrity of the estimated position is assessed.
However, addressing state estimation and integrity monitoring separately does not capture the uncertainty in the state estimation algorithm.
As a result, the integrity monitoring becomes biased by the acquired  state estimate leading to subsequent faulty state estimation.

Recently, an approach towards joint state estimation and integrity monitoring for GNSS measurements was proposed in Particle RAIM \cite{p}.
Instead of producing point positioning estimates, Particle RAIM uses a particle filter to form a multi-modal probability distribution over position, represented as particles.
Traditional RAIM \cite{raim} is used to assess the correctness of different ranging measurements and the particle weights are updated to form the distribution over the position.
From the resulting probability distribution, the integrity risk is derived using an approximate upper bound to the probability of HMI or the reference risk.
By incorporating the correctness of different measurement subsets directly into the state estimation, Particle RAIM is able to exclude multiple faults in GNSS ranging measurements.
However, due to large errors from GNSS measurements, Particle RAIM requires employing conservative measures such as large Alert Limits to adequately bound the reference risk. 

For urban applications, improved positioning accuracy from Particle RAIM is necessary to provide adequate integrity for smaller Alert Limits.
Since measurements from GNSS are not sufficient to provide the desired accuracy, it is helpful to augment GNSS with additional sensors that increase redundancy in measurements.
Sensors such as cameras are effective complimentary sensors to GNSS. 
In urban regions, cameras have access to rich environmental features \cite{Ram} \cite{c1} \cite{c2} and provide superior sensing than GNSS which suffers from  multi-path and NLOS errors \cite{trad} \cite{c3} \cite{c4} \cite{c5}. 

Thus, with added vision, we need  a framework to provide integrity for the fused GNSS-camera navigation system to account for two categories of faults. The first category includes data association errors across images, where repetitive features are found in multiple images creating ambiguity during feature and image association.
This ambiguity is further amplified due to variations in lighting and environmental conditions.
The second category comprises errors that arise during sensor fusion of GNSS and camera measurements. 
Ensuring that faults in either measurement do not dominate the sensor fusion process  is paramount for maximizing the complimentary characteristics of GNSS and camera.

Many works provide integrity for GNSS-camera fused systems utilizing a Kalman Filter \cite{kf} framework or an information filter \cite{if}. Vision Aided-RAIM \cite{va} introduced landmarks as pseudo-satellites and integrated them into a linear measurement model alongside GPS observations.
In \cite{seq}, the authors implemented a sequential integrity monitoring approach to isolate single satellite faults. 
The integrity monitor uses the innovation sequence output from a single Kalman filter to derive a recursive expression of the worst case failure mode slopes and to compute the protection levels (PL) in real-time. 
An Information Filter (IF) is used in \cite{wkl} for data fusion wherein faults are detected based on the Kullback-Leibler divergence (KL divergence) \cite{kl} between the predicted and the updated distributions. 
After all detected faulty measurements are removed, the errors are modeled by a student’s t distribution to compute a PL.
A student's t distribution is also used in \cite{tdist} alongside informational sensor fusion for fault detection and exclusion.
The degree of the distribution is adapted in real-time based on the computed residual from the information filter.
A distributed information filter is proposed in \cite{dif} to detect faults in GPS measurement by checking the consistency through log-likelihood ratio of the information innovation of each satellite.
These approaches model  measurement fault distributions with a Gaussian distribution although for camera measurements, the true distribution might be non-linear, multi-modal, and arbitrary in nature. 
Using a simplified linear measurement probability distribution renders these frameworks infeasible and unreliable for safety-critical vision augmented GNSS applications.

Another line of work builds on Simultaneous Localization and Mapping (SLAM) based factor graph optimization techniques. 
Bhamidipati \textit{et al}  \cite{Ram} derived PL by modeling GPS satellites as global landmarks and introducing image pixels from a fish-eye camera as additional landmarks. The raw image is categorized into sky and non-sky pixels to further distinguish between LOS and NLOS satellites. 
The overall state is estimated using graph optimization along with an M-estimator.
Although this framework is able to exclude multiple faults in GPS measurements, it is not extendable to measurements from forward or rear facing cameras that do not capture sky regions.
Along similar lines, measurements from a stereo camera along with GNSS pseudoranges are jointly optimized in a graph optimization framework in \cite{Gong}. GNSS satellites are considered as feature vision points and pose-graph SLAM is applied to achieve a positioning solution.
However, graph optimization approaches also share the same limitation as Kalman Filter based approaches: They produce point positioning estimates and do not account for the uncertainty in state estimation that biases integrity monitoring.

Overall, existing integrity monitoring algorithms for GNSS- camera fusion have the following limitations:
\begin{itemize}
\item[1] They address state estimation and integrity monitoring separately, similar to traditional RAIM approaches. 
\item[2] They accommodate camera measurements within a linear or linearizable framework such as KF, EKF, or IF and become infeasible when camera measurements are not linearizable without loss of generality.
\item[3] There is no standard way in literature to quantify the uncertainty in camera measurements directly from raw images.
\item[4] They use outlier rejection techniques to perform fault detection and exclusion after obtaining the positioning solution. 
There is no framework that accounts for faults both independently in GNSS and camera as well as the faults that arise during sensor fusion. 
\end{itemize} 

In our work, we overcome the above limitations by proposing the following contributions. This paper is based on our recent ION GNSS+ 2020 conference paper \cite{me}. 

\begin{itemize}
\item[1] 	We jointly address state estimation and integrity monitoring for GNSS-camera fusion with a particle filtering framework. We retain the advantages of Particle RAIM while extending it to include camera measurements.
 \item[2] We derive a probability distribution over position directly from images leveraging image registration.
 \item[3] We develop a metric based on KL divergence \cite{kl} to fuse probability distributions obtained from GNSS and camera measurements. 
By minimizing the KL divergence  of the distribution from each camera measurement with respect to the GNSS measurement distribution, we ensure that erroneous camera measurements do not affect the overall probability distribution. Stated otherwise, the divergence metric augments the shared belief over the position from both sensor measurements by minimizing cross-contamination during sensor fusion.
\item[4] We experimentally validate our framework on an urban environment dataset \cite{ds} with faults in GNSS and camera measurements.
\end{itemize}

The rest of the paper is organized as follows. In Section 2, we describe the overall particle filter framework for probabilistic sensor fusion. In Sections 3 and 4, we infer a distribution over position from GNSS and camera measurements, respectively.  Section 5 elaborates on the probabilistic sensor fusion of GNSS and camera measurements along with the proposed KL divergence metric. In Section 6, we describe the integrity risk bounding. Sections 7 and 8 shows the experimental setup and the results from experimental validation on the urban environment dataset, respectively. In Section 9, we conclude our work.

\section{PARTICLE FILTER FRAMEWORK FOR PROBABILISTIC SENSOR FUSION}
\begin{figure}[H]
\begin{center}
 \includegraphics[scale=0.6]{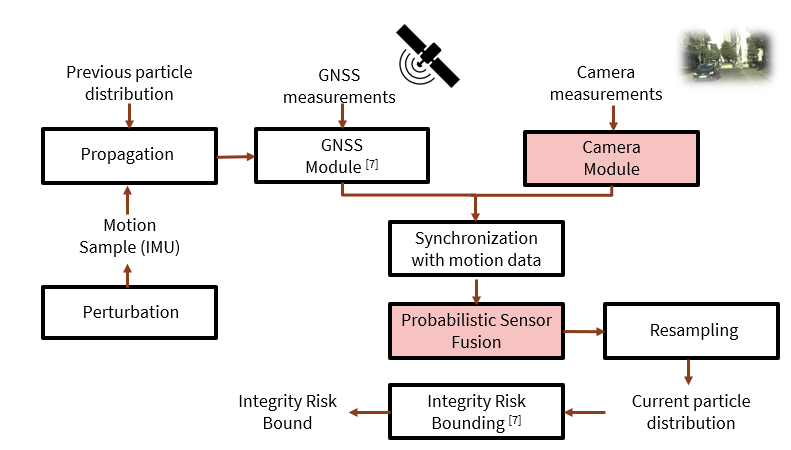} 
  \end{center}
  \caption{Particle filter framework with probabilistic sensor fusion of GNSS and camera measurements and integrity risk bounding. The highlighted modules represent our contributions.The GNSS and Risk Bounding Modules are adopted from Particle RAIM \cite{p}.}
  \label{fig:ovf}
\end{figure}

The distribution over the position inferred from GNSS and camera measurements is multi-modal due to faults in a subset of measurements. Thus, to model such distributions, we choose a particle filtering approach that further allows us to keep track of multiple position hypotheses rather than a single position estimate.
Although a particle filtering approach was used in Particle RAIM \cite{p}, the authors only considered GNSS ranging measurements.
In our work, we extend the framework to include measurements from a camera sensor.
Figure \ref{fig:ovf} represents our overall framework. We add the camera and probabilistic sensor fusion modules to the framework proposed in \cite{p}.
Our framework consists of the following modules:
\begin{itemize}
\item[•] Perturbation and propagation: Using noisy inertial odometry from the IMU, we generate a set of motion samples, each of which perturbs the previous particle distribution in the propagation step. 
\item[•] GNSS module: This module from Particle RAIM \cite{p} takes GNSS ranging measurements from multiple satellites, some of which may be faulty and outputs a probability distribution over position using a fault-tolerant weighting scheme described in Section~\ref{section:gnss}.
The particles from the GNSS module are propagated to the camera module to ensure that the distributions from GNSS and camera share the same domain of candidate positions.
\item[•]Camera module and synchronization with motion data: The camera module takes a camera image and matches it to the images in a map database using image registration to generate similarity scores.
The underlying state of the best matched image is extracted and propagated forward to the current GNSS time stamp by interpolating with IMU odometry.
This step ensures that the probability distributions from camera and GNSS measurements are generated at the same time stamps. 
Finally, we use a categorical distribution function to transform the similarity scores into a probability distribution over position hypotheses as described in Section 4.
\item[•]Probabilistic sensor fusion: This module outputs a joint likelihood over positions from GNSS and camera measurements after fusing them with the proposed KL divergence metric in Section \ref{section:klm}. Particles are resampled from the current distribution with Sequential Importance Resampling \cite{pf2}.
\item[•]Risk bounding: We adopt the risk bounding formulation proposed in \cite{p} to compute the integrity risk from the derived probability distribution over the position domain. Using generalization bounds from statistical learning theory \cite{st}, the derived risk bound is formally shown to over bound the reference risk in  Section \ref{section:irisk}.
\end{itemize}

We elaborate on the various modules of our framework in the following sections.

\section{GNSS MODULE- PARTICLE RAIM}
\label{section:gnss}
\begin{figure}[H]
\begin{center}
\includegraphics[scale=0.4]{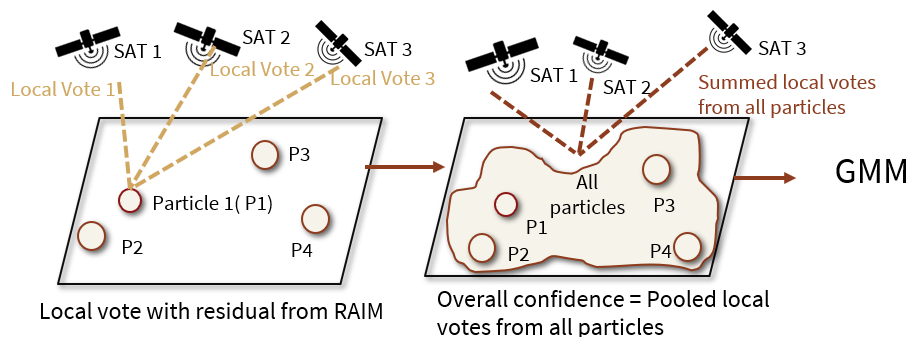} 
  \end{center}
  \caption{Two steps of the EM scheme used to derive the weight of each Gaussian likelihood in the GMM. In the expectation step, the local vote for each particle is computed based on the squared-normal voting on the normalized residual for a particle obtained with traditional RAIM. The overall confidence is inferred by normalizing the votes and pooling them using Bayesian maximum \textit{a posteriori} (MAP) estimation.}
  \label{fig:praim}
\end{figure}

A likelihood model for the GNSS measurements is derived using the mixture weighting method proposed in Particle RAIM \cite{p}.
Instead of assuming correctness of all GNSS measurements, the likelihood is modeled as a mixture of Gaussians to account for faults in some measurements.
Individual measurement likelihoods are modeled as Gaussians with the expected pseudoranges as means and variance based on Dilution of Precision(DoP). 
The GMM \cite{sor} \cite {sor2} is expressed as:
\begin{equation}
L_{t} (m^{t}) = \sum_{k=0}^{R} \gamma_{k} \mathcal{N} (m_{k}^{t} | \mu_{X}^{t,k} , \sigma_{X}^{t,k}) ; \sum_{k=0}^{R} \gamma_{k} = 1,
\end{equation}
where $L_{t} (m^{t})   $ denotes the likelihood of measurement $ m $ at time $ t $. $ \gamma $ denotes the measurement responsibility or the weights of the individual measurement components and $R$ refers to the total number of GNSS ranging measurements.
$ \mu $ and $ \sigma $ represent the mean and the standard deviation of each Gaussian component inferred from DOP. $ X $ refers to the collection of position hypotheses denoted by particles and $ k $ is the index of the number of Gaussians in the mixture.
The weights are inferred with a single step of the Expectation-Maximization (EM) scheme \cite{em} as shown in Figure \ref{fig:praim}.

To avoid numerical errors due to finite precision, the log likelihood of the likelihood model is implemented by extending the input space to include additional copies of the state space variable, one for each GNSS measurement \cite{b}. 
The new likelihood is written as:
\begin{equation}
P\left(m^t\middle| X^t,\chi=k\right)=\gamma_k\mathcal{N}\left(m_k^t\middle|\mu_x^{t,k},\sigma_x^{t,k}\right)\ ; \sum_{k=1}^{R}\gamma_k=1,
\end{equation}
where $ \chi $ is an index that denotes the associated GNSS measurement with the particle replica.

\section{CAMERA MODULE}
\begin{figure}[H]
\begin{center}
  \includegraphics[scale=0.6]{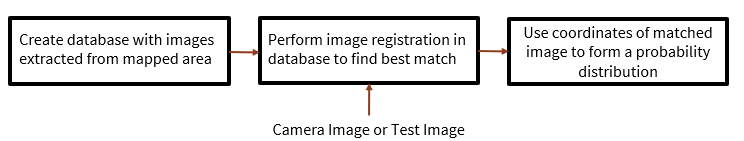} 
  \end{center}
  \caption{ Generating probability distribution over position from camera images.}
  \label{fig:cm}
\end{figure}

To quantify the uncertainty from camera images, we use a map-matching algorithm that matches a camera image directly to an image present in a map database.
Our method is implemented in OpenCV \cite{ocv} and comprises three steps shown in Figure \ref{fig:cm}.

Each block is elaborated below.
\begin{itemize}
\item[•]Database Creation: We assume prior knowledge of the geographical region where we are navigating. Based on GPS coordinates, we select images from the known area using Google Street View Imagery. 
These images along with their associated coordinates form the database. Features are extracted from these images and stored in a key point-descriptor format.
\item[•]Image Registration: After receiving a camera test image, we extract features and descriptors with the ORB \cite{orb} algorithm. 
Although we experimented with other feature extraction methods such as SIFT \cite{sift}, SURF \cite{surf}, and AKAZE \cite{aka}, ORB was found most effective for extracting descriptors from highly blurred images.
The descriptor vectors are clustered with a k-means algorithm \cite{kmeans} to form a vocabulary tree \cite{svt}.
Each node in the tree corresponds to an inverted file, i.e., a file containing the ID-numbers of images in which a particular node is found and the relevance of each feature to that image.
The database is then scored hierarchically based on Term Frequency Inverse Document Frequency (TF-IDF) scoring \cite{svt}, which quantifies the relevance of the images in the database to the camera image. 
We refer to these scores as the similarity scores.
The image with the highest score is chosen as the best match and the underlying state is extracted. 

\item[•]Probability generation after synchronization: 
After extracting the state from the best camera image in the database, we propagate the state to the same time stamp as the GNSS measurement. 
The raw vehicle odometry is first synchronized with GNSS measurements using the algorithm in \cite{ds}.
Using the time difference between the previous and current GNSS measurements, we linearly interpolate the extracted state with IMU motion data as shown below.
\begin{equation} \label{eqn}
x^{t} = x^{t-1} + v^{t-1} dt +  0.5 a^{t-1}\ dt^{2}
\end{equation}
where $x^{t} $ refers to the 3D position at epoch $t  $, $ dt $ refers to the time difference between successive camera measurements, and $ v $ and $ a $ are the interpolated IMU velocity and accelerations at epoch $ t $.

Next, we compute the Euclidean distance between the interpolated state and the current particle distribution from GNSS measurements to obtain new similarity scores. 
This step ensures that the probability distributions computed from camera and GNSS measurements share the same domain of candidate positions.
A SoftMax function takes the scores and outputs a probability distribution over position. Normalization of the scores enforces a unit integral for the distribution.
\begin{equation}
Q (n^{t}| X^{t}) = \frac{\exp(\omega^t)}{\sum_c \exp(\omega_c^t)} 
\end{equation}
where
$ Q $ is the probability distribution associated with camera measurement $ n $ at time $ t $ over the position domain $X $,
$ \omega_{c}^{t} $ represents computed distance score, and $ c $ is the index for individual particles.
\end{itemize}

\section{PROBABILISTIC SENSOR FUSION}
\label{section:psf}
After obtaining the probability distributions from GNSS and camera, we need to form a joint distribution over the position.
However, we need to ensure that faults in camera measurements do not degrade the distribution from GNSS measurements, one that is coarse but correct since the distribution accounts for faults in the ranging measurements through the RAIM voting scheme.
Thus, we need a metric to identify and exclude faulty camera measurements leveraging knowledge of the distribution from GNSS.
Additionally, the metric should assess the consistency of the probability distribution from each camera measurement with respect to the GNSS distribution and mitigate inconsistent distributions that result from vision faults.
The KL divergence \cite{kldd2} represents one way to assess the consistency of two probability distributions. 
By minimizing the divergence between the distributions inferred from camera and GNSS, we ensure that both distributions are consistent.

\subsection{Kl Divergence: Metric Formulation}
We provide a background on KL divergence prior to explaining our metric.

\label{section:klm}
The KL divergence \cite{kldd2} between two discrete probability distributions, $ p $ and $ q $, in the same domain is defined as:
\begin{equation}
D_{KL} ( p || q) = \sum\nolimits_{z \in \zeta} p _{z} \ log \ \frac{p _{z}}{q_{z}}
\end{equation}

where $ \zeta $ represents the domain of both distributions and $ z $ is each element of the domain. In our work, we ensure that distributions from GNSS and camera share the same position domain by propagating the particles from the GNSS distribution to the camera module prior to generating the distribution from camera measurements.
Two important properties of the KL divergence are:
\begin{itemize}
\item[•] The KL divergence between two distributions is always non-negative and not symmetrical \cite{kldd2}
\begin{equation}
D_{KL}( p || q) \neq D_{KL} ( q || p) 
\end{equation}
where $D_{KL} ( q || p)$ is the reverse KL divergence between the distributions $ p $ and $ q $.
\item[•]$D_{KL} ( p || q)  $ is convex in the pair $  ( p || q) $ if both distributions represent probability mass functions (pmf) \cite{kldd2}.
\end{itemize}

Leveraging the above properties, we formulate our metric below.
\begin{itemize}
\item[•] Mixture of Experts (MoE): 
We form a mixture distribution to represent probability distributions from successive camera measurements, where a non-Gaussian probability distribution is derived from a single camera image.
Each measurement is assigned a weight to represent its contribution in the mixture.
Instead of setting arbitrary weights, we leverage the GNSS distribution to infer weights that directly correspond to whether a camera measurement is correct or faulty.
Thus, highly faulty camera measurements are automatically assigned low weights in the MoE. 
The mixture distribution is given as:
\begin{equation}
Q^{*} (n^{t} | X^{t}) = \sum\limits_{j=1}^K \alpha_j^* \ Q^{j}  (n_{j}^{t} | X^{t}); 
 \sum\limits_{j=1}^K \alpha_j^*  = 1
\end{equation}
where $Q^{*} (n^{t} | X^{t})  $ represents the mixture distribution formed using $K$ camera images between two successive GNSS time epochs. $ Q^{j}  (n_{j}^{t} | X^{t}) $ is the likelihood of a single camera image $ n_{j}^{t} $ recorded at time $t $ with $\alpha_j^* $ as the normalized weight. $ X^{t} $ are the particles representing position hypothesis and $j $ is the index for the camera images.
The weights are normalized below to ensure that the MoE forms a valid probability distribution:
\begin{equation}
\alpha_j^* = \frac{\alpha_j}{\sum\limits_{r=1}^K \alpha_r}
\end{equation}
where
$\alpha_j^* $ is the normalized weight, $\alpha_j$ is the weight prior to normalization, $r $ is the index for the number of camera images between two successive GNSS time epochs, and $K $ is the total number of camera measurements. 

\item[•]Setup KL divergence: We set up a divergence minimization metric between the distributions from each camera measurement and all GNSS measurements.
\begin{equation}
{KL}_j\ ((\alpha_j\ Q^{j}\left(n_j^t\middle| X^t\right)\ ||\ P\ \left(m_k^t\middle| X^t,\chi=k\right))=\sum_{i=1}^{S}{\left(\alpha_j\ Q^j(n_j^t|\ X^t)\right)\ log\left[\frac{\left(\alpha_j\ Q^j(n_j^t|\ X^t)\right)}{P\ \left(m_k^t\middle| X^t,\chi=k\right)}\right]}
\end{equation}
where $||\   $ denotes the divergence between both probability distributions, $ S $  represents the total number of particles or position hypotheses across both distributions, and $ i $ is the index for the particles. 
$\ P\ \left(m_k^t\middle| X^t,\chi=k\ \right) $ is the probability distribution at epoch $ t $ from GNSS measurements as defined in Equation (2), 
$ \alpha_j\ $ is the unnormalized weight, and $j $ is the index for the camera measurement.

\item[•] Minimize divergence: Using the convexity of the KL divergence (Property 2), we minimize each divergence metric with respect to the unknown weight assigned to the likelihood of each camera measurement.
We abbreviate $\ P\left(m_i^t\middle| X^t,\chi=i\ \right) $  as $P (x_{i}) $ and $Q\left(n_j^t\middle| X^t\right)\ $ as $Q(x_{i}) $
for brevity and expand Equation (9). Since $ \alpha_{j} $ is independent of the summation index, we keep it outside the summation and simplify our expansion below. 
\begin{equation}
{KL}_j(Q|\left|P\right)=\ \alpha_j\sum_{i\ =\ 1}^{S}Q\left(x_i\right)log\ \alpha_j\ +\ \alpha_j\sum_{i\ =\ 1}^{S}Q\left(x_i\right)log\ Q\left(x_i\right)\ -\ \alpha_j\sum_{i\ =\ 1}^{S}Q\left(x_i\right)log\ P\ \left(x_i\right)
\end{equation}
Taking the first derivative with respect to $ \alpha_{j} $ we obtain,
\begin{equation}
{min}_{\alpha_j}{KL}_j\ \ (Q|\left|P\right)=\ log\ \alpha_j\sum_{i\ =\ 1}^{S}Q\left(x_i\right)\ +\ \sum_{i\ =\ 1}^{S}Q\left(x_i\right)\ +\sum_{i\ =\ 1}^{S}Q\left(x_i\right)log\ Q\left(x_i\right)\ -\ \sum_{i\ =\ 1}^{S}Q\left(x_i\right)log\ P\left(x_i\right)
\end{equation}

Equating the expression on the right to 0 and solving for $\alpha_j $ gives us:
\begin{equation}
\alpha_j=e^k\ ;\ k=\frac{\sum_{i=1}^{S}Q\left(x_i\right)\ log\ \frac{P\left(x_i\right)}{Q\left(x_i\right)}}{\sum_{i=1}^{S}Q\left(x_i\right)}-1
\end{equation}

We also perform a second derivative test to ensure that the $ \alpha_{j} $ value inferred is a minimum value of the divergence measure. Since the exponential function  with the natural base is always positive,  $ \alpha_{j} $ is always positive as well.
Thus, evaluating the second derivative gives us a positive value.
\begin{equation}
\frac{1}{\alpha_j}\sum_{i=1}^{S} Q (x_i) > 0
\end{equation}

\item[•]
Joint probability distribution over position:
After obtaining the weights, we normalize them using Equation (8). 
We obtain the joint distribution assuming that the mixture distribution from camera measurements and the GMM from GNSS measurements are mutually independent.
The joint distribution  is given as:
\begin{equation}
P^\ast\left(n^t,\ m^t\middle| X^t\right)=\ P \left(m_i^t\middle| X^t,\chi=k\ \right)\ Q^\ast\left(n^t\middle| X^t\right)
\end{equation}
where $\ P\ \left(m_k^t\middle| X^t,\chi=k\ \right) $
is the probability distribution from GNSS measurements in Equation (2).
We take the log likelihood of the joint distribution to avoid finite precision errors. 
\end{itemize}

\section{INTEGRITY RISK BOUNDING}
\label{section:irisk}
We upper bound the probability of HMI using the risk bounding framework introduced in \cite{p}.
For a single epoch, the probability of HMI for a given Alert Limit $ r $ is defined as:
\begin{equation}
R_{x*} (\pi) = \mathop{\mathbb{E}}_{x\sim \pi}[ P (\norm{x - x^{*}}\geq  r)]
\end{equation}
where
$R_{x*} (\pi) $ is the probability of HMI with reference position $ x^{*} $ and mean distribution in position space induced by all posterior distributions $ \pi $.
The distributions are created by generating samples around the measured odometry and then perturbing the initial particle distribution.
From the PAC-Bayesian \cite{pac} formulation and as shown in \cite{p}, the reference risk $\mathop{\mathbb{\textbf{R}}( \pi^{t})}$ upper bound is:
\begin{equation}
\mathop{\mathbb{\textbf{R}}( \pi^{t})} \leq
\mathop{\mathbb{\textbf{R}}_M ( \pi^t)} + \mathcal{D}_{Ber}^{-1} (\mathop{\mathbb{\textbf{R}}_M ( \pi^t)}, \epsilon)
\end{equation}
 The first and second terms refer to empirical and divergence risk, respectively. We explain the computation of each term below.
 
 The empirical risk  $\mathop{\mathbb{\textbf{R}}_M ( \pi^{t})}$ is computed from a finite set of perturbed samples of size $ M $.
 \begin{equation}
\mathop{\mathbb{\textbf{R}}_M ( \pi^{t})} =  \frac{1}{M} \sum_{i=1}^{M}\mathop{\mathbb{E}}_{x\sim \pi^{t}} [l (x, \pi_u^{t})],
\end{equation}
where,  $l (x, \pi_u^{t}) $ is the classification loss with respect to a motion sample resulting in the posterior distribution being classified as hazardous. $\pi$ refers to the mean posterior distribution at time $t$.

The divergence risk term  $\mathcal{D}_{Ber}^{-1} (\mathop{\mathbb{\textbf{R}}_M ( \pi^{t})}, \epsilon)$ accounts for uncertainty due to perturbations that are not sampled.
First, we compute the gap term  $ \epsilon$  using KL divergence \cite{kl} of the current distribution from the prior and a confidence requirement in the bound $\delta$.
\begin{equation}
\epsilon = \frac{1}{M} ( KL (\pi^t || \pi^{t-1}) + log (\frac{M +1}{\delta}))
\end{equation}
where $ \delta $ refers to the bound gap. The means of the prior and current distributions are taken as $ \pi^{t-1} $ and $ \pi^{t} $. The prior and current distributions are approximated as multivariate Gaussian distributions.

The Inverse Bernoulli Divergence \cite{p} $\mathcal{D}_{Ber}^{-1}$ is defined as:
\begin{equation}
\mathcal{D}_{Ber}^{-1}(q, \epsilon) = t \; \; s.t. \; \;   \mathcal{D}_{Ber}( q || q + t) = \epsilon
\end{equation}
where $q || q+ t $ is the KL divergence \cite{kl} between $q$ and $q + t $ and $q $ is given by the empirical risk term.
Finally, the Inverse Bernoulli Divergence \cite{p} is obtained approximately as:
\begin{equation}
\mathcal{D}_{Ber}(q, \epsilon) = \sqrt{\frac{2 \epsilon}{\frac{1}{q} + \frac{1}{1- q}}}
\end{equation}

\section{EXPERIMENTS}
\subsection{Datasets}
We test our framework on a 2.3 km long urban driving dataset from Frankfurt \cite{ds}.
We use GNSS pseudorange measurements, images from a forward-facing camera, ground truth from a NovAtel receiver, and odometry from the IMU.
The dataset contains NLOS errors in GNSS measurements and vision faults due to variations in illumination.
In addition to the real-world dataset, we create emulated datasets by inducing faults in GNSS and vision measurements with various controlled parameters. 
 
\subsection{Experimental Setup and Parameters}
\begin{itemize}
\item[•]Real-world dataset: We use GNSS ranging measurements with NLOS errors. For simplicity, we estimate the shared clock bias by subtracting the average residuals with respect to ground truth from all GNSS pseudoranges at one time epoch. 
\item[•]Emulated dataset: First, we vary the number of satellites with NLOS errors by adding back the residuals to randomly selected satellites. This induces clock errors in some measurements which are perceived as faults. 
Secondly, we remove the NLOS errors from all measurements but add Gaussian bias noise to pseudorange measurements from random satellites at random time instances. The number of faults are varied between 2-9 out of 12 available measurements at any given time step. We induce faults in camera measurements by adding blurring with a  21x21 Gaussian kernel and occlusions of 25-50 \% height and width to random images.
\end{itemize}

During the experimental simulation, a particle filter tracks the 3D position (x,y,z) of the car and uses faulty GNSS and camera measurements along with noisy odometry.
Probability distributions are generated independently from GNSS and camera and fused with the KL divergence metric to form the joint distribution over positions.
At each time epoch, the particle distribution with the highest total log-likelihood is chosen as the estimated distribution for that epoch. 
The integrity risk is computed from 10 posterior distributions of the initial particle distribution and the reference risk is computed with ground truth.
Our experimental parameters are listed in Table 1.  

\begin{table}[ht]
\centering
\caption{Experimental Parameters for Validation with Real-world and Emulated Datasets}
 \begin{tabular}{||c c c c||} 
 \hline
 Parameter & Value &  Parameter &  Value \\ [0.5ex] 
 \hline\hline
No. of GNSS measurements & 12 & Added Gaussian bias to GNSS measurements & 20- 200 m \\
 \hline
No. of faults in GNSS measurements & 2-9 & No. of particles & 120 \\
 \hline
Measurement noise variance & 10 m $^{2}$ & Filter propagation variance & 3 m $^{2}$ \\
 \hline
Alert Limit & 8, 16 m	 & No. of odometry perturbations & 10 \\ 
[1ex] 
 \hline
\end{tabular}
\end{table}

\subsection{Baselines and Metrics}
We use Particle RAIM as the baseline to evaluate our algorithm's performance for state estimation. 
The metric for state estimation is the root mean square error (RMSE) of the estimated position with respect to ground truth for the entire trajectory.
The risk bounding performance is evaluated with metrics derived from a failure event, i.e., when the derived risk bound fails to upper bound the reference risk. 
The metrics are the following: failure ratio(the fraction of cases where the derived risk bound fails to upper bound the reference risk), failure error(the mean error during all failure events), and the bound gap(average gap between the derived integrity risk) and the reference risk. 

For evaluating the integrity risk, we specify a performance requirement that the position should lie within the Alert Limit with at least 90\% probability. 
A fault occurs if the positioning error exceeds the Alert Limit.
The metrics for integrity risk are reported based on when the system has insufficient integrity or sufficient integrity \cite{peso}, which respectively refer to the states when a fault is declared or not.
The false alarm rate equals the fraction of the number of times the system declares insufficient integrity in the absence of a fault.
The missed identification rate is defined as the fraction of the number of times the system declares sufficient integrity even though a fault is present.

\section{RESULTS}
\subsection{State Estimation}
First, we test our algorithm with NLOS errors in GNSS ranging measurements and added camera faults. 
Quantitative results in Table 2 demonstrate that our algorithm produces 3D positioning estimates with overall RMSE of less than 11 m.
Additionally, our algorithm reports lower errors compared to Particle RAIM for all test cases.
Our algorithm is able to compensate for the residual errors from Particle RAIM by including camera measurements in the framework.
This leads to improved accuracy in the positioning solution.

\begin{table}[ht]
\centering
\caption{RMSE in 3D Position with NLOS errors and added vision faults}
\begin{tabular}{|c|c|c|c|c|c|}
\hline 
No. of faults out of 12 available GNSS measurements & Particle RAIM-Baseline (meter)  & Our Algorithm (meter) \\ 
\hline 
2 &  18.1& 6.3 \\ 
\hline 
4 &  19.1 &  6.1\\ 
\hline 
6 & 16.9 & 5.9\\ 
\hline 
9 &  26.6 & 10.6 \\ 
\hline 
\end{tabular} 
\end{table}

For qualitative comparison, we overlay the trajectories from our algorithm on ground truth and highlight regions with positioning error of greater than 10 m in Figures 4 and 5.
Trajectories from Particle RAIM show large deviations from ground truth in certain regions, either due to poor satellite signal availability or high NLOS errors in the faulty pseudorange measurements.
However, similar deviations are absent from the trajectories from our algorithm which uses both GNSS and camera measurements. 
Our KL divergence metric is able to mitigate the errors from vision and the errors from cross-contamination during sensor fusion, allowing us to produce lower positioning error.
\begin{figure}[H]
\centering
\begin{subfigure}{.5\textwidth}
  \centering
\includegraphics[scale=0.75]{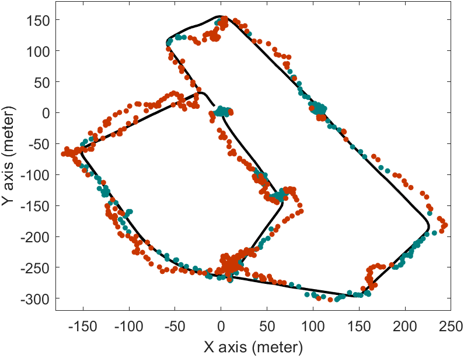} 
  \caption{Particle RAIM (Baseline)}
  \label{fig:sub1}
\end{subfigure}%
\begin{subfigure}{.5\textwidth}
  \centering
 \includegraphics[scale= 0.75]{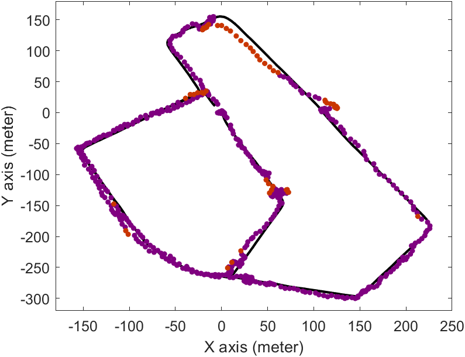} 
  \caption{Our Algorithm}
  \label{fig:sub2}
\end{subfigure}
\label{fig:test}
\caption{State estimation under NLOS errors for 6 faulty GNSS pseudo range measurements and added vision faults. Regions with positioning error greater than 10 m are highlighted in red.}
\end{figure}

\begin{figure}[H]
\centering
\begin{subfigure}{.5\textwidth}
  \centering
\includegraphics[scale=0.75]{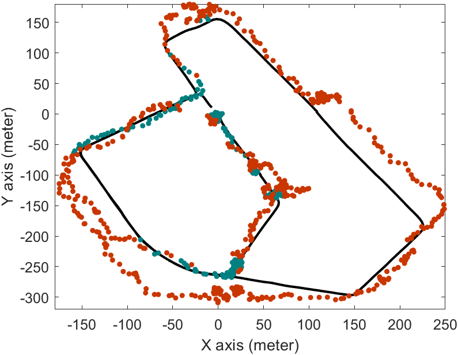} 
  \caption{Particle RAIM (Baseline)}
  \label{fig:sub1}
\end{subfigure}%
\begin{subfigure}{.5\textwidth}
  \centering
\includegraphics[scale=0.75]{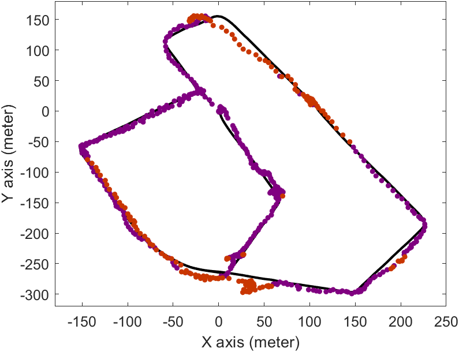} 
  \caption{Our Algorithm}
  \label{fig:sub2}
\end{subfigure}
\label{fig:test}
\caption{State estimation under NLOS errors for 9 faulty GNSS pseudo range measurements and added vision faults. Regions with positioning error greater than 10 m are highlighted in red.}
\end{figure}

Secondly, we test our algorithm with the emulated datasets.
Quantitatively, we plot the RMSE as a function of the added Gaussian bias value in Figure 6 and as a function of the number of faulty GNSS ranging measurements in Figure 7.
For all validation cases, our algorithm produces an overall RMSE less than 10 m.
Similar to the results from the real-world dataset, our algorithm reports lower RMSE values than Particle RAIM.
With a fixed number of faults, the errors generally increase with increasing bias.
At a fixed bias value, the errors decrease with increasing number of faults up to 6 faulty GNSS measurements since large number of faults are easily excluded by Particle RAIM producing an improved distribution over the position.
The improved distribution from GNSS further enables the KL divergence metric to exclude faulty camera measurements and produce a tighter distribution over the position domain.
However, with a higher number of faults, Particle RAIM does not have enough redundant correct GNSS measurements to exclude the faulty measurements resulting in higher positioning error.
Nevertheless, with added vision, our algorithm produces better positioning estimates for all test cases than Particle RAIM.

\begin{figure}[H]
\centering
\begin{subfigure}{.3\textwidth}
  \centering
\includegraphics[scale=0.5]{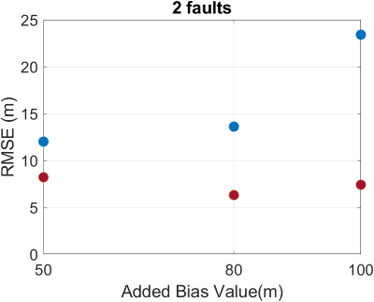} 
  \label{fig:sub1}
\end{subfigure}%
\begin{subfigure}{.3\textwidth}
  \centering
\includegraphics[scale=0.5]{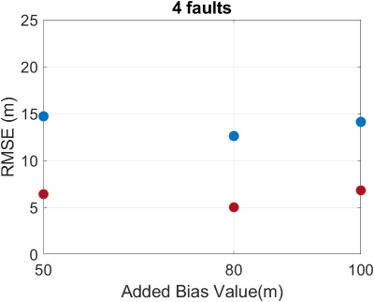} 
  \label{fig:sub1}
\end{subfigure}%
\begin{subfigure}{.3\textwidth}
  \centering
\includegraphics[scale=0.5]{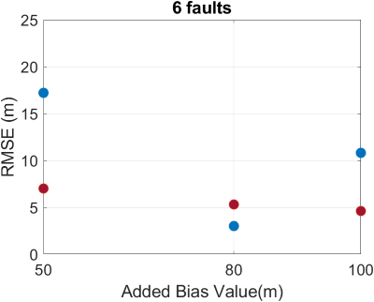} 
  \label{fig:sub2}
\end{subfigure}
\newline
\begin{subfigure}{.3\textwidth}
\includegraphics[scale=0.7]{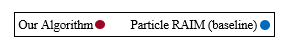} 
 \label{fig:sub1}
\end{subfigure}%
\caption{RMSE from our algorithm and Particle RAIM (baseline) for varying numbers of faults in GNSS ranging measurements at a fixed added Gaussian bias value.}
\label{fig:test}
\end{figure}

\begin{figure}[H]
\centering
\begin{subfigure}{.3\textwidth}
  \centering
\includegraphics[scale=0.5]{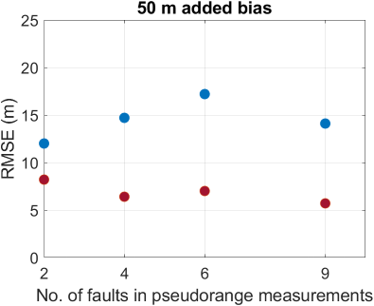} 
  \label{fig:sub1}
\end{subfigure}%
\begin{subfigure}{.3\textwidth}
  \centering
\includegraphics[scale=0.5]{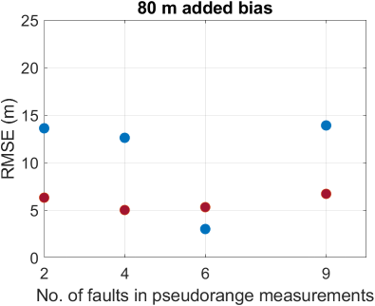} 
  \label{fig:sub1}
\end{subfigure}%
\begin{subfigure}{.3\textwidth}
  \centering
\includegraphics[scale=0.5]{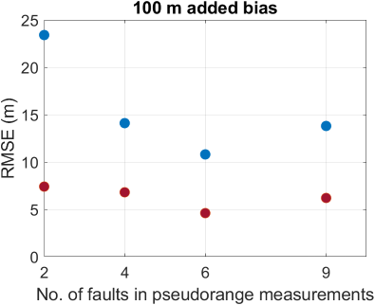} 
  \label{fig:sub2}
\end{subfigure}
\newline
\begin{subfigure}{.3\textwidth}
\includegraphics[scale=0.7]{leg.PNG} 
 \label{fig:sub1}
\end{subfigure}%
\caption{RMSE from our algorithm and Particle RAIM (baseline) for various added Gaussian bias values with fixed number of faulty GNSS measurements.}
\label{fig:test}
\end{figure}

\subsection{Integrity Monitoring}

\begin{figure}[H]
\centering
\begin{subfigure}{.3\textwidth}
  \centering
\includegraphics[scale=0.5]{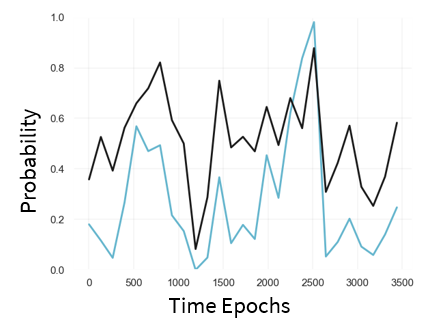} 
  \label{fig:sub1}
\end{subfigure}%
\begin{subfigure}{.3\textwidth}
  \centering
\includegraphics[scale=0.5]{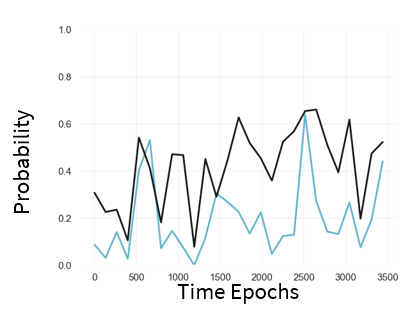}  
  \label{fig:sub1}
\end{subfigure}%
\begin{subfigure}{.3\textwidth}
  \centering
 \includegraphics[scale=0.5]{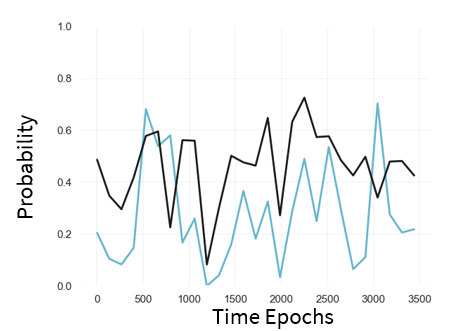} 
  \label{fig:sub2}
\end{subfigure}
\newline
\begin{subfigure}{.3\textwidth}
\includegraphics[scale=0.6]{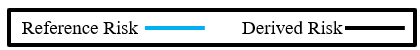} 
 \label{fig:sub1}
\end{subfigure}%
\caption{Reference risk and integrity risk bound with 8 m Alert Limit for varying numbers of faults and added bias of 100 m in GNSS measurements. The derived risk bound over bounds the reference risk with less than 0.11 failure ratio for all test cases.  }
\label{fig:im1}
\end{figure}

\begin{figure}[H]
\centering
\begin{subfigure}{.3\textwidth}
  \centering
\includegraphics[scale=0.5]{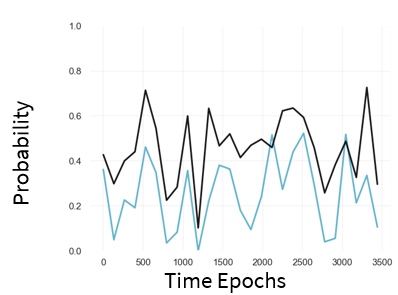} 
  \label{fig:sub1}
\end{subfigure}%
\begin{subfigure}{.3\textwidth}
  \centering
\includegraphics[scale=0.5]{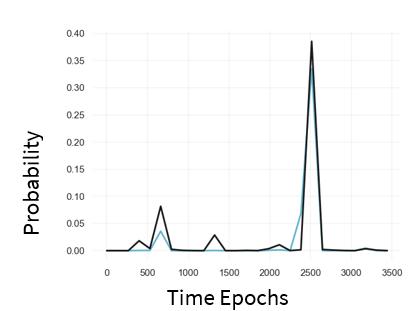}  
  \label{fig:sub1}
\end{subfigure}%
\begin{subfigure}{.3\textwidth}
  \centering
 \includegraphics[scale=0.5]{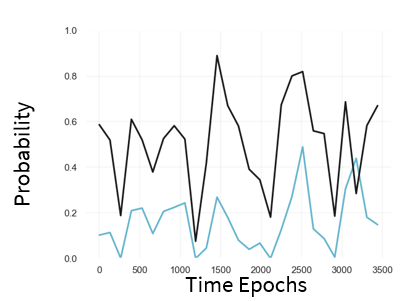} 
  \label{fig:sub2}
\end{subfigure}
\newline
\begin{subfigure}{.3\textwidth}
\includegraphics[scale=0.6]{leg2.PNG} 
 \label{fig:sub1}
\end{subfigure}%
\caption{Reference risk and integrity risk bound with 8 m Alert Limit for varying numbers of faults and added bias of 200 m in GNSS measurements. The derived risk bound over bounds the reference risk with less than 0.07 failure ratio for all test cases. }
\label{fig:im2}
\end{figure}

We evaluate the integrity risk bounding performance for two Alert Limits, 8 m and 16 m. 
For an Alert Limit of 8 m, Table \ref{table:8m} shows that the derived integrity risk satisfies the performance requirement with very low false alarm and missed identification rates.
While the false alarm rates reported are 0 for all test cases except two and the missed identification rates are always less than 0.11. 
Additionally, the integrity risk bound upper bounds the reference risk with a failure ratio of less than 0.11 and a bound gap of less than 0.4 for all cases.
Figures \ref{fig:im1} and \ref{fig:im2} further support the observation that the derived risk bound is able to over bound the reference risk with low failure rate for the same Alert Limit.
The few instances when the derived risk bound fails to upper bound the reference risk occur due to large sudden jumps in the reference risk that go undetected considering the fixed size of our motion samples.
However, in general, the integrity risk produced from our algorithm is able to satisfy the desired performance requirement and successfully overbound the reference risk for an Alert Limit as small as 8 m.
This choice of Alert Limit is allowed because of the low positioning errors that further enable non-conservative integrity risk bounds.

\begin{table}[ht]
\centering
\caption{Integrity Risk for Alert Limit of 8 m }
\label{table:8m}
\begin{tabular}{|c|c|c|c|c|c|c|}
\hline 
Added Bias Value (meter)  & No. of Faults & $ P_{FA} $ & $ P_{MI} $ & Failure Ratio & Failure Error (meter) & Bound Gap\\ 
\hline 
100 & 2  & 0 & 0.03 & 0.07 & 7.5 & 0.26 \\ 
\hline 
100 & 4  & 0 & 0.04 & 0.04 & 2.3 & 0.25 \\ 
\hline 
100 & 6  & 0 & 0.07 & 0.11 & 2.9 & 0.25 \\  
\hline 
100 & 9  & 0.07 & 0.03 & 0.07 & 4.7 & 0.36 \\ 
\hline 
200 & 2  & 0 & 0.07 & 0.07 & 3.5 & 0.20 \\ 
\hline 
200 & 4  & 0.11 & 0 & 0.04 & 4.8 & 0.40 \\ 
\hline 
200 & 6  & 0 & 0 & 0 & - & 0.38 \\ 
\hline 
200 & 9  & 0 & 0.07 & 0.04 & 5.4 & 0.36 \\ 
\hline 
\end{tabular} 
\end{table}

For an Alert Limit of 16 m, Table \ref{table:16m} shows that the integrity risk satisfies the integrity performance requirement with 0 false alarm rates.
Furthermore, the missed identification rates are always 0 except for the test case with 9 faults and 100 m added bias.
Specifying a larger Alert Limit lowers the risk associated with the distribution over position since almost all particles from the perturbed distributions lie within the Alert Limit. Thus, the integrity risk with a 16 m Alert Limit is reported to be much smaller compared to the risk obtained with a 8 m Alert Limit as shown in Figures \ref{fig:im1} and \ref{fig:im2}. 
Additionally, the derived risk bound produces even lower failure ratio of less than 0.07 and a tighter bound gap of less than 0.1.
Overall, the derived risk bound over bounds the reference risk for various bias and fault scenarios in Figures 10 and 11.

\begin{table}[ht]
\centering
\caption{Integrity Risk for Alert Limit of 16 m}
\label{table:16m}
\begin{tabular}{|c|c|c|c|c|c|c|}
\hline 
Added Bias Value (meter)  & No. of Faults & $ P_{FA} $ & $ P_{MI} $ & Failure Ratio & Failure Error (meter) & Bound Gap\\ 
\hline 
100 & 2  & 0 & 0 & 0 & - & 0.10 \\ 
\hline 
100 & 4  & 0 & 0 & 0 & - & 0.08 \\ 
\hline 
100 & 6  & 0 & 0 & 0.04 & 5.9 & 0.05 \\  
\hline 
100 & 9  & 0 & 0.04 & 0.07 & 9.7 & 0.08 \\ 
\hline 
200 & 2  & 0 & 0 & 0.07 & 5.0 & 0.09 \\ 
\hline 
200 & 4  & 0 & 0 & 0.07 & 4.2 & 0.07 \\ 
\hline 
200 & 6  & 0 & 0 & 0 & 3.6 & 0.06 \\ 
\hline 
200 & 9  & 0 & 0 & 0.04 & 3.8 & 0.01 \\ 
\hline 
\end{tabular} 
\end{table}

\begin{figure}[H]
\centering
\begin{subfigure}{.3\textwidth}
  \centering
\includegraphics[scale=0.5]{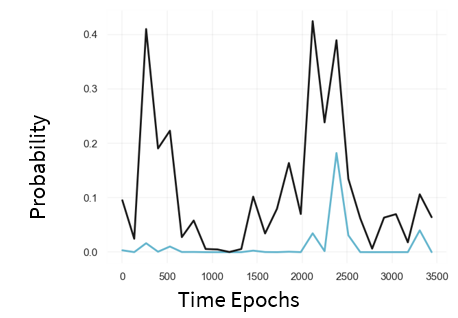} 
  \label{fig:sub1}
\end{subfigure}%
\begin{subfigure}{.3\textwidth}
  \centering
\includegraphics[scale=0.5]{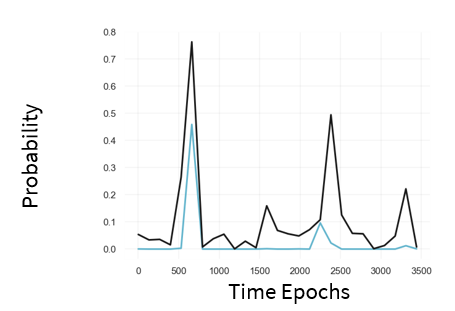}  
  \label{fig:sub1}
\end{subfigure}%
\begin{subfigure}{.3\textwidth}
  \centering
 \includegraphics[scale=0.5]{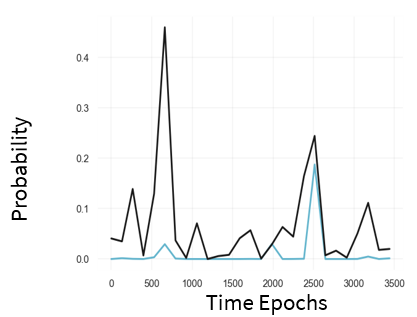} 
  \label{fig:sub2}
\end{subfigure}
\newline
\begin{subfigure}{.3\textwidth}
\includegraphics[scale=0.6]{leg2.PNG} 
 \label{fig:sub1}
\end{subfigure}%
\caption{Reference risk and integrity risk bound with 16 m Alert Limit for varying numbers of faults and added bias of 100 m in GNSS measurements. The derived risk bound over bounds the reference risk with less than 0.07 failure ratio for all test cases. }
\label{fig:im3}
\end{figure}

\begin{figure}[H]
\centering
\begin{subfigure}{.3\textwidth}
  \centering
\includegraphics[scale=0.5]{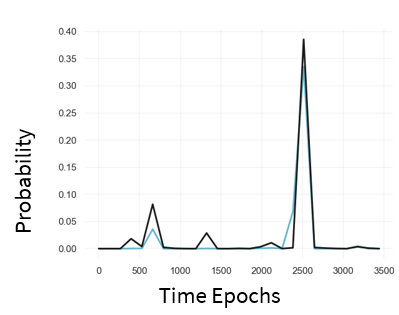} 
  \label{fig:sub1}
\end{subfigure}%
\begin{subfigure}{.3\textwidth}
  \centering
\includegraphics[scale=0.5]{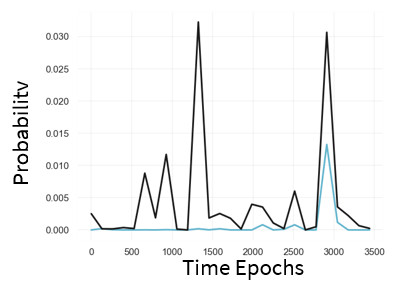}  
  \label{fig:sub1}
\end{subfigure}%
\begin{subfigure}{.3\textwidth}
  \centering
 \includegraphics[scale=0.5]{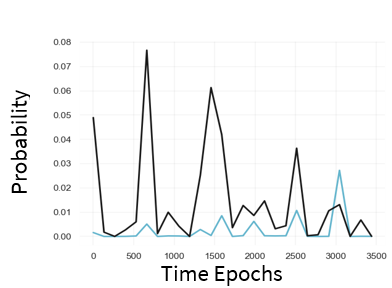} 
  \label{fig:sub2}
\end{subfigure}
\newline
\begin{subfigure}{.3\textwidth}
\includegraphics[scale=0.6]{leg2.PNG} 
 \label{fig:sub1}
\end{subfigure}%
\caption{Reference risk and integrity risk bound with 16 m Alert Limit for varying numbers of faults and added bias of 200 m GNSS measurements. The derived risk bound over bounds the reference risk with less than 0.07 failure ratio for all test cases. }
\label{fig:im4}
\end{figure}

\section{CONCLUSION}
In this paper, we presented a framework for joint state estimation and integrity monitoring for a GNSS-camera fused system using a particle filtering approach.
To quantify the uncertainty in camera measurements, we derived a probability distribution directly from camera images leveraging a data-driven approach along with image registration.
Furthermore, we designed a metric based on KL divergence to probabilistically fuse measurements from GNSS and camera in a fault-tolerant manner. 
The metric accounts for vision faults and mitigates the errors that arise due to cross-contamination of measurements during sensor fusion.
We experimentally validated our framework on real-world data under NLOS errors, added Gaussian bias noise to GNSS measurements, and added vision faults.
Our algorithm reported lower positioning error compared to Particle RAIM which uses only GNSS measurements.
The integrity risk from our algorithm satisfied the integrity performance requirement for Alert Limits of 8 m and 16 m with low false alarm and missed identification rates.
Additionally, the derived integrity risk successfully provided an upper bound to the reference risk with a low failure rate for both Alert Limits, making our algorithm suitable for practical applications in urban environments.

\section{ACKNOWLEDGMENT}
We express our gratitude to Akshay Shetty, Tara Mina and other members of the Navigation of Autonomous Vehicles Lab for their feedback on early drafts of the paper.

\bibliographystyle{ieeetr}
\bibliography{mybib}

\begin{thebibliography}{10}

\bibitem{p}
S.~Gupta and G.~X. Gao, ``Particle raim for integrity monitoring,'' {\em
  Proceedings of the 32nd International Technical Meeting of the Satellite
  Division of The Institute of Navigation, ION GNSS+ 2019}, 2019.

\bibitem{kl}
H.~Zhu, ``On information and sufficiency,'' 04 1997.

\bibitem{trad}
N.~{Zhu}, J.~{Marais}, D.~{Bétaille}, and M.~{Berbineau}, ``Gnss position
  integrity in urban environments: A review of literature,'' {\em IEEE
  Transactions on Intelligent Transportation Systems}, vol.~19, no.~9,
  pp.~2762--2778, 2018.

\bibitem{raim}
Y.~C. Lee, ``Analysis of range and position comparison methods as a means to
  provide gps integrity in the user receiver,'' {\em Proceedings of the 32nd
  International Technical Meeting of the Satellite Division of The Institute of
  Navigation)}, vol.~19, no.~9, pp.~1--4, 1986.

\bibitem{Ram}
S.~Bhamidipati and G.~Gao, ``Slam-based integrity monitoring using gps and
  fish-eye camera,'' {\em Proceedings of the 32nd International Technical
  Meeting of the Satellite Division of The Institute of Navigation, ION GNSS+
  2019}, pp.~4116--4129, 10 2019.

\bibitem{c1}
Z.~{Wang}, Y.~{Wu}, and Q.~{Niu}, ``Multi-sensor fusion in automated driving: A
  survey,'' {\em IEEE Access}, vol.~8, pp.~2847--2868, 2020.

\bibitem{c2}
J.~{Rife}, ``Collaborative vision-integrated pseudorange error removal:
  Team-estimated differential gnss corrections with no stationary reference
  receiver,'' {\em IEEE Transactions on Intelligent Transportation Systems},
  vol.~13, no.~1, pp.~15--24, 2012.

\bibitem{c3}
{He Chengyan}, {Guo Ji}, {Lu Xiaochun}, and {Lu Jun}, ``Multipath performance
  analysis of gnss navigation signals,'' pp.~379--382, 2014.

\bibitem{c4}
S.~M. {Steven Miller}, X.~{Zhang}, and A.~{Spanias}, {\em Multipath Effects in
  GPS Receivers: A Primer}.
\newblock 2015.

\bibitem{c5}
K.~{Ali}, X.~{Chen}, F.~{Dovis}, D.~{De Castro}, and A.~J. {Fernández}, ``Gnss
  signal multipath error characterization in urban environments using lidar
  data aiding,'' pp.~1--5, 2012.

\bibitem{kf}
R.~E. Kalman, ``A new approach to linear filtering and prediction problems,''
  {\em Transactions of the ASME--Journal of Basic Engineering}, vol.~82,
  no.~Series D, pp.~35--45, 1960.

\bibitem{if}
X.~Wang, N.~Cui, and J.~Guo, ``Information filtering and its application to
  relative navigation,'' {\em Aircraft Engineering and Aerospace Technology},
  vol.~81, pp.~439--444, 09 2009.

\bibitem{va}
L.~Fu, J.~Zhang, R.~Li, X.~Cao, and J.~Wang, ``Vision-aided raim: A new method
  for gps integrity monitoring in approach and landing phase,'' {\em Sensors
  (Basel, Switzerland)}, vol.~15, pp.~22854--73, 09 2015.

\bibitem{seq}
C.~Tanil, S.~Khanafseh, M.~Joerger, and B.~Pervan, ``Sequential integrity
  monitoring for kalman filter innovations-based detectors,'' {\em Proceedings
  of the 32nd International Technical Meeting of the Satellite Division of The
  Institute of Navigation, ION GNSS+ 2018}, 10 2018.

\bibitem{wkl}
J.~{Al Hage} and M.~E. {El Najjar}, ``Improved outdoor localization based on
  weighted kullback-leibler divergence for measurements diagnosis,'' {\em IEEE
  Intelligent Transportation Systems Magazine}, pp.~1--1, 2018.

\bibitem{tdist}
J.~Al~Hage, P.~Xu, and P.~Bonnifait, ``Bounding localization errors with
  student’s distributions for road vehicles,'' {\em International Technical
  Symposium on Navigation and timing}, 11 2018.

\bibitem{dif}
N.~A. {Tmazirte}, M.~E.~E. {Najjar}, C.~{Smaili}, and D.~{Pomorski},
  ``Multi-sensor data fusion based on information theory. application to gnss
  positionning and integrity monitoring,'' {\em 15th International Conference
  on Information Fusion}, pp.~743--749, 2012.

\bibitem{Gong}
Z.~Gong, P.~Liu, Q.~Liu, R.~Miao, and R.~Ying, ``Tightly coupled gnss with
  stereo camera navigation using graph optimization,'' {\em Proceedings of the
  32nd International Technical Meeting of the Satellite Division of The
  Institute of Navigation, ION GNSS+ 2018}, pp.~3070--3077, 10 2018.

\bibitem{me}
A.~{Mohanty}, S.~{Gupta}, and G.~X.{Gao}, ``A particle filtering framework for
  integrity risk of gnss-camera sensor fusion,'' {\em Proceedings of the 33 nd
  International Technical Meeting of the Satellite Division of The Institute of
  Navigatin, ION GNSS+ 2020}, 2020.

\bibitem{ds}
P.~Reisdorf, T.~Pfeifer, J.~Bre{\ss}ler, S.~Bauer, P.~Weissig, S.~Lange,
  G.~Wanielik, and P.~Protzel, ``The problem of comparable gnss results – an
  approach for a uniform dataset with low-cost and reference data,'' in {\em
  The Fifth International Conference on Advances in Vehicular Systems,
  Technologies and Applications} (M.~Ullmann and K.~El-Khatib, eds.), vol.~5,
  p.~8, nov 2016.
\newblock ISSN: 2327-2058.

\bibitem{pf2}
F.~{Gustafsson}, F.~{Gunnarsson}, N.~{Bergman}, U.~{Forssell}, J.~{Jansson},
  R.~{Karlsson}, and P.~. {Nordlund}, ``Particle filters for positioning,
  navigation, and tracking,'' {\em IEEE Transactions on Signal Processing},
  vol.~50, no.~2, pp.~425--437, 2002.

\bibitem{st}
S.~B. O.~Bousquet and G.~Lugosi, ``Introduction to statistical learning
  theory,'' {\em Advanced Lectures on Machine Learning}, vol.~3176,
  pp.~169--207, 2004.

\bibitem{sor}
M.~Simandl and J.~Dunik, ``Design of derivative-free smoothers and
  predictors,'' {\em 14th IFAC Symposium on System Identification, Newcastle,
  Australia}, 03 2006.

\bibitem{sor2}
H.~W. Sorenson and D.~L. Alspach, ``Recursive bayesian estimation using
  gaussian sums,'' {\em Automatica}, 1971.

\bibitem{em}
J.~P. {Vila} and P.~{Schniter}, ``Expectation-maximization gaussian-mixture
  approximate message passing,'' {\em IEEE Transactions on Signal Processing},
  vol.~61, no.~19, pp.~4658--4672, 2013.

\bibitem{b}
C.~M. Bishop, {\em {Pattern recognition and machine learning}}.
\newblock Information science and statistics, New York, NY: Springer, "2006".

\bibitem{ocv}
G.~Bradski, ``The opencv library,'' {\em Dr. Dobb's Journal of Software Tools},
  2000.

\bibitem{orb}
E.~{Rublee}, V.~{Rabaud}, K.~{Konolige}, and G.~{Bradski}, ``Orb: An efficient
  alternative to sift or surf,'' {\em International Conference on Computer
  Vision}, pp.~2564--2571, 2011.

\bibitem{sift}
D.~Lowe, ``Distinctive image features from scale-invariant keypoints,'' {\em
  International Journal of Computer Vision}, vol.~60, pp.~91--110, 11 2004.

\bibitem{surf}
H.~Bay, T.~Tuytelaars, and L.~Van~Gool, ``Surf: Speeded up robust features,''
  {\em European Conference on Computer Vision}, vol.~3951, pp.~404--417, 07
  2006.

\bibitem{aka}
P.~F. Alcantarilla, J.~Nuevo, and A.~Bartoli, ``Fast explicit diffusion for
  accelerated features in nonlinear scale spaces,'' {\em British Machine Vision
  Conference}, 2013.

\bibitem{kmeans}
S.~P. Lloyd, ``Least squares quantization in pcm,'' {\em Information Theory,
  IEEE Transactions}, vol.~28.2, pp.~129--137, 1982.

\bibitem{svt}
D.~{Nister} and H.~{Stewenius}, ``Scalable recognition with a vocabulary
  tree,'' {\em IEEE Computer Society Conference on Computer Vision and Pattern
  Recognition}, vol.~2, pp.~2161--2168, 2006.

\bibitem{kldd2}
T.~Erven and P.~Harremoës, ``Rényi divergence and kullback-leibler
  divergence,'' {\em Information Theory, IEEE Transactions on}, vol.~60,
  pp.~3797--3820, 2014.

\bibitem{pac}
L.~G. Valiant, ``A theory of the learnable,'' {\em Commun. ACM}, vol.~27,
  p.~1134–1142, Nov. 1984.

\bibitem{peso}
H.~Pesonen, ``A framework for bayesian receiver autonomous integrity monitoring
  in urban navigation,'' {\em NAVIGATION: Journal of the Institute of
  Navigation}, vol.~58, pp.~229--240, 09 2011.

\end{thebibliography}

\end{document}